\pdfoutput=1

\documentclass[11pt]{article}

\usepackage{emnlp2021}

\usepackage{caption}
\usepackage{subcaption}
\usepackage{times}
\usepackage{latexsym}
\usepackage{microtype}
\usepackage{multirow}
\usepackage{float}
\usepackage{amsmath}
\usepackage{graphicx}
\usepackage{tabularx}
\usepackage{booktabs}
\usepackage[final]{changes}
\usepackage{pbox}
\usepackage{latexsym}
\usepackage{microtype}
\usepackage{amssymb}
\usepackage{caption}
\usepackage{array}
\usepackage{makecell}
\usepackage{array}
\usepackage{changes}
\usepackage{dblfloatfix}
\usepackage{pgfplots}
\pgfplotsset{width=7cm,compat=1.8}
\usepackage{colortbl}
\usepackage{pgfplotstable}
\usepackage{listings}
\usepackage{ stmaryrd }
\usepackage{stmaryrd}
\usepackage{trimclip}
\usepackage{hyperref}
\usepackage{enumitem}
\usepackage{soul}

\makeatletter
\DeclareRobustCommand{\shortto}{%
  \mathrel{\mathpalette\short@to\relax}%
}

\newcommand{\short@to}[2]{%
  \mkern2mu
  \clipbox{{.3\width} 0 0 0}{$\m@th#1\vphantom{+}{\shortrightarrow}$}%
  }
\makeatother

\definecolor{lightgray}{rgb}{0.83, 0.83, 0.83}
\definecolor{purple}{rgb}{0.5,0,1}
\definecolor{dcyan}{rgb}{0.2,0.6,0.5}
\definecolor{darkgreen}{rgb}{0,200,0}
\definecolor{light-gray}{gray}{0.95} 
\definecolor{darkgreen}{RGB}{0,140,0}
\definecolor{lightblue}{RGB}{199,226,255}
\definecolor{lightorange}{RGB}{250,175,50}

\newcommand{\orangetext}[1]{\colorbox{lightorange}{#1}}

\usepackage[T1]{fontenc}

\usepackage[utf8]{inputenc}
\usepackage{enumitem}

\newcommand{\subscript}[2]{$#1 _ #2$}

\usepackage{microtype}

\ifthenelse{\equal{1}{2}}
{ 
\newcommand{\daniel}[1]{\textcolor{cyan}{[DK:#1]}}
\newcommand{\changed}[1]{\textcolor{purple}{#1}}
\newcommand{\nascomment}[1] {\textcolor{red}{\textsc{\textbf{[#1 --Noah]}}}}
\newcommand{\suchin}[1] {\textcolor{blue}{\textsc{\textbf{[#1 --suchin]}}}}
\newcommand{\karishma}[1] {\textcolor{green}{\textsc{\textbf{[#1 --karishma]}}}}
\newcommand{\kelvin}[1] {\textcolor{dcyan}{\textsc{\textbf{[#1 --kellu]}}}}
\newcommand{\changedKel}[1]{\textcolor{dcyan}{#1}}
\newcommand{\delete}[1] {\textcolor{red}{\textsc{\textbf{[#1 ]}}}}

}
{
\newcommand{\daniel}[1]{}
\newcommand{\changed}[1]{}
\newcommand{\nascomment}[1] {}
\newcommand{\suchin}[1] {}
\newcommand{\karishma}[1] {}
\newcommand{\kelvin}[1] {}
\newcommand{\changedKel}[1]{}
\newcommand{\delete}[1] {}

}
\usepackage{xspace}
\newcommand{\temporaldrift}{temporal misalignment\xspace}

\newcommand{\github}{https://github.com/Kel-Lu/time-waits-for-no-one}

\newcommand{\timestamsps}{\mathcal{T}}

\title{\emph{Time Waits for No One!} \\ A Study of Factors Contributing to Temporal Drift}
\title{\emph{Time Waits for No One!} \\ An Analysis on Temporal Drift of Language Models}
\title{Temporal Drift of Language Models: Analysis and Challenges}
\title{
 \vspace*{-0.5in}
{{\small \hfill NAACL 2022}\\
 \vspace*{.25in}} 
Time Waits for No One! \\ Analysis and Challenges of Temporal Misalignment}

\newcommand{\andd}{\hspace{2em}}
\newcommand{\aitwo}{$^2$}
\newcommand{\uw}{$^1$}

\author{
 Kelvin Luu\uw \andd Daniel Khashabi\aitwo \andd Suchin  Gururangan\uw \\  \textbf{Karishma Mandyam\uw} \andd
\textbf{Noah A. Smith}\uw$^,$\aitwo \vspace{.35em}\\
\uw University of Washington \quad
\aitwo Allen Institute for AI \quad
\\
\small
\texttt{\{kellu,sg01,krm28,nasmith\}@cs.washington.edu}, \\
\small \texttt{danielk@allenai.org}
}

\begin{document}
\maketitle
\begin{abstract}
When an NLP model is trained on text data from one time period and tested or deployed on data from another, the resulting \emph{\temporaldrift} can degrade end-task performance.
In this work, we establish a suite of eight diverse tasks across different domains (social media, science papers, news, and reviews) and periods of time (spanning five years or more) to quantify the effects of \temporaldrift.
Our study is focused on the ubiquitous setting where a pretrained model is optionally adapted through continued domain-specific pretraining, followed by task-specific finetuning. We establish a suite of tasks across multiple domains to study \temporaldrift in modern NLP systems.
We find stronger effects of \temporaldrift 
on task performance than have been previously reported.
We  also find that, while temporal adaptation through continued pretraining can help, these gains are small compared to task-specific finetuning on data from the target time period.  
Our findings motivate continued research to improve temporal robustness of NLP models.
\footnote{\href{\github}{Data and code are available here.}}
\end{abstract}

\section{Introduction}
Changes in the ways a language is used over time are widely attested
\cite{Labov1994PrinciplesOL,altmann2009beyond,eisenstein2014diffusion}; how these changes will affect NLP systems built from text corpora, and in particular their long-term performance, is not as well understood.

This paper focuses on \emph{\temporaldrift}, i.e., where training and evaluation datasets are drawn from different periods of time.  In today's pretraining-finetuning paradigm, this misalignment can affect a pretrained language model---a situation that has received recent attention \citep{jaidka2018diachronic,lazaridou2021pitfalls,peters2018deep,raffel2020exploring, rottger2021temporal}---or the finetuned task model, or both.  We suspect that the effects of \temporaldrift will vary depending on the genre or domain of the task's text, the nature of that task or application, and the specific time periods.

We focus primarily on measuring the extent of \temporaldrift on task performance.  We consider eight tasks, each with datasets that span at least five years (\S\ref{subsec:datasets}), ranging from summarization to entity typing, a subproblem of entity recognition \citep{Grishman1999AME}.  Notably, these task datasets span four different domains:  social media, scientific articles, news, and reviews.  We introduce an easily interpretable metric that summarizes the rate at which task performance degrades as function of time.

Our research questions are:
\begin{enumerate}[noitemsep, label=(\subscript{Q}{{\arabic*}}),leftmargin=28pt]
    \item \label{Q1} \emph{how does \temporaldrift{} affect downstream tasks over  time?} 
    \item \label{Q2} \emph{how does sensitivity to \temporaldrift{} vary with text domain and task? }
    \item \label{Q3} \emph{how does \temporaldrift{} affect language models across domains and spans of time?}
    \item \label{Q4} \emph{how effective is temporal adaptation, or additional pretraining on a target year, in mitigating \temporaldrift{}?}
\end{enumerate}

We find that \temporaldrift{} affects both language model generalization and task performance. We find considerable variation in degradation across text domains (\S\ref{sec:domain:adaptation}) and tasks (\S\ref{sec:downstream:tasks}). Over five years, classifiers' $F_1$ score can deteriorate as much as 40 points (political affiliation in Twitter) or as little as 1 point (Yelp review ratings).  Two distinct tasks defined on the same domain can show different levels of degradation over time.

We explore domain adaptation of a language model, using temporally selected (unannotated) data, as a  way to curtail  \temporaldrift{}  \citep{rottger2021temporal}. 
We find that this does not offer much benefit, especially relative to performance that  can be achieved by finetuning on temporally suitable data (i.e., from the same time period as the test data). We conclude that temporal adaptation should not be seen as a substitute for finding temporally aligned labeled data.

The evidence and benchmarks we offer motivate careful attention to \temporaldrift in many applications of NLP models, and further research on solutions to this problem.

\paragraph{Contributions.} 
To facilitate the study of \temporaldrift{} phenomenon on downstream applications, we compile a suite of eight diverse tasks across four important language domains.
We define an interpretable metric that summarizes \temporaldrift{} of a model on a task with timestamped data.  Our experiments reveal key factors in how \temporaldrift affects NLP model performance.



\section{Methodology Overview}
\label{sec:setup}

We begin by defining the scope of our study.

\subsection{Learning Pipeline}
\label{subsec:model:pipeline}

\begin{figure}[h]
    \centering
    \includegraphics[scale=0.76,trim=0.5cm 0.6cm 0cm 0.1cm]{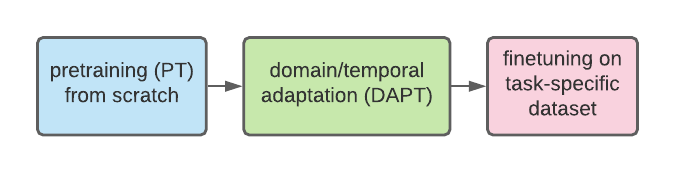}
    \caption{A typical modeling pipeline in NLP. }
    \label{fig:pipeline}
\end{figure}

We consider a process for building an NLP model that is in widespread use by the research community, illustrated in Fig.~\ref{fig:pipeline}.  First, a (neural network) language model  (LM) is pretrained on a large text collection that is not necessarily selected for topical or temporal proximity to the text of the target application (our focus is on GPT-2; \citealp{brown2020language}).  Second, the LM is optionally adapted by continued training on a collection strategically curated for closer proximity to the target \citep{beltagy2019scibert}; this stage is often referred to as domain-adaptive pretraining (DAPT; \citealp{gururangan2020don}).  Finally, the model is finetuned to minimize a task-specific loss, using labeled data representative of what the model is expected to be exposed to in testing or deployment.

We study two ways in which \temporaldrift might affect the pipeline's performance as well as straightforward ways to mitigate them.




\paragraph{Task Shift and Temporal Finetuning} The relationship between text inputs and target outputs may change over time.  To the extent that this occurs, annotated datasets used to train NLP systems in the finetuning stage will  become stale over time. Due to this \temporaldrift, performance will degrade after deployment, or any in evaluations that use test data temporally distant from the training data. We seek to quantify this degradation across a range of text domains and tasks.

\paragraph{Language Shift and Temporal Domain Adaptation}
Changes in language use can cause a pretrained LM, which commonly serves as the backbone for most modern NLP models, to become stale over time \citep{lazaridou2021pitfalls}, regardless of the end task.  \citet{lazaridou2021pitfalls} explored \emph{temporal adaptation}, continuing LM training on new text data.  This is essentially the same procedure as DAPT, where the data is selected by time period. Their work focused on the LM alone, not downstream tasks; we consider both here. 

\citet{rottger2021temporal}, the closest to our work, studied temporal adaptation in conjunction to finetuning for a classification task over Reddit data. They conclude that temporal adaptation does not help any more than normal DAPT. We corroborate this work and extend it by studying a wider variety of tasks over a longer span of time periods and thus are better able to draw generalizations from our results. 

We believe that the two kinds of shift---task shift and language shift---are difficult to disentangle, and we do not attempt to do so in this work.  Instead, we aim to quantify the effect of \temporaldrift on a range of NLP tasks, as well as the benefits of these two strategies.

\subsection{Evaluation Methodology}

Our experiments are designed to measure the effect of \temporaldrift on task performance.  To do so, for each task, we fix a test set within a given time period, $T_{\mathit{test}}$.  We vary the time period of the training data, allowing us to interpret differences in performance as a kind of ``regret'' relative to the performance of a model trained on data temporally aligned with $T_{\mathit{test}}$.\footnote{This setup avoids a confound of varying test set difficulty that we would encounter if we fixed the model and compared its performance across  test datasets from \emph{different} time periods.} We consider multiple different test periods for each task. 
We also seek to control the effect of training dataset size.  We partition training data into time periods of roughly the same size and always train on a single partition, keeping the training set size of each time period constant within each task.  We expect that  performance could be improved by accumulating training data across multiple time periods, but that would make it more difficult to achieve our research goal of quantifying the effect of \temporaldrift on performance.

\subsection{Quantifying Temporal Degradation}
\label{subsec:metric}

Understanding \temporaldrift requires evaluating a model's performance across data with a range of different timestamps, which makes it difficult to compare various models in terms of their misalignment. 
We define a metric for temporal degradation (TD) which summarizes  \added{the expected speed of model degradation due to} \temporaldrift on a task as a single value. In high-level terms, the TD score measures the average rate of performance deterioration (of perplexity, $F_1$, or Rouge-L) \added{for each timestep of difference between that the train and evaluation sets}. Higher TD scores imply greater levels of performance deterioration due to misalignment.

\begin{figure}
\centering
\includegraphics[trim=0.5cm 0.5cm 0cm 0cm, width=.35\textwidth]{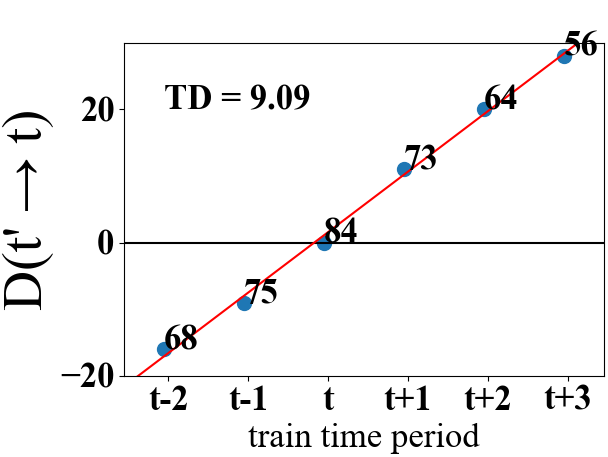}
\caption{
    An example calculation of the TD score for a particular timestep $t$ (discussed in Section~\ref{subsec:metric}).
        The plotted markers represent $D(t'\to t)$ ($y$-axis) as a function of train time period $t'$  ($x$-axis). 
        The annotated numbers on each blue dot are the raw evaluation scores $S_{t'\to t}$, not to be confused with the $y$ values.
    The red line is the line of best fit and its slope is the TD score for evaluation timestep $t$. In this example, we would expect to see, on average, $9.09$ points of deterioration for each year of misalignment. The final TD score is averaged across all evaluation timesteps. 
}
\label{figure:td_example}
\end{figure}

Let $S_{t' \shortto t}$ indicate the performance a model trained on timestep $t'$ data and evaluated on timestep $t$. We define ${D} (t' \shortto{} t)$ as:
$$
{D} (t' \shortto{} t) = 
    -\left(S_{t' \shortto t} - S_{t \shortto t} \right) \times \text{sign}(t' - t).
$$
${D} (t' \shortto{} t)$ is a modified difference in performance between two models.\footnote{Without the modification, a task with degradation would have have positive performance gaps both $t' > t$ and $t' > t$; the function would not be monotone and the rate of change would be harder to approximate. The modification yields a simpler visual understanding of the deviations over time.} Fig.~\ref{figure:td_example} illustrates $D$ as a function of consecutive training time periods. 


We find a line of best fit for ${D} (t' \shortto{} t)$ for all $t'$ using least-squares regression. The slope of this line is $\text{TD}(t)$, the TD score for evaluation time period $t$. The final TD score is the average of the $\text{TD}(t)$ across all evaluation time periods $t$. Further details can be found in Appendix~\ref{appendix:subsec:metric}.

\subsection{Domains, Tasks, and  Datasets}
\label{subsec:datasets}

\begin{table*}[t]
    \small 
    \centering
    \resizebox{\textwidth}{!}{%
    \begin{tabular}{ccccm{10cm}}
        \toprule
            Domain & Task & Time Range & Size & Example \\ 
        \midrule
            \multirow{3}{*}{Twitter}  & \makecell{political\\affiliation\\classification} & 2015-2019 & $120k$ &  \textbf{Input:} \emph{History will note that Trump didn’t merely fiddle while the planet burned but tried to throw the Arctic National W... } \textbf{Output:} \emph{Democrat (vs Republican) }  \\ 
            \cline{2-5}
            & \makecell{entity type\\classification} & 2014-2019 & $8k$  &  \textbf{Input:} \emph{entity: Finola, tweet: Two 64-year olds enjoying their first birthday together in 40+ years. My twin sister, Finola, and I.}    \textbf{Output:} \emph{Person}  \\ 
            \specialrule{.001em}{.001em}{.001em} 
            \specialrule{.001em}{.001em}{.001em} 
            \multirow{4}{*}{Science} & \makecell{mention\\type\\classification} & 1980-2016 & $8k$ &   \textbf{Input:}  \emph{mention: deep Long Short-Term Memory (LSTM) subnetwork, abstract: In this paper, we study the problem of online action detection from the streaming skeleton data .... by leveraging the merits of the deep Long Short-Term Memory (LSTM) subnetwork, the proposed model ...} \textbf{Output:} \emph{Method}  \\
            \cline{2-5}
             & \makecell{venue\\classification} & 2009-2020 & $16k$ & \textbf{Input:} \emph{Rank K Binary Matrix Factorization (BMF) approximates a binary matrix by the product of two binary matrices of lower rank, K...}  \textbf{Output:} \emph{AAAI (vs ICML)} \\ 
            \specialrule{.001em}{.001em}{.001em} 
            \multirow{7}{*}{News} & \makecell{media \\ frame\\classification} & 2009-2016 & $20k$ & \textbf{Input:} \emph{You think you have heard the worst horror a gun in the wrong hands can do, and then this.You think there could not have been anywhere more tragic for it to happen...}  \textbf{Output:} \emph{Gun Control (15 possible frames)} \\ 
            \cline{2-5}
             & \makecell{publisher \\ classification} & 2009-2016 & 67$k$ & \textbf{Input:} \emph{A Muslim woman said Sunday that her viral article explaining why she voted for Donald Trump has angered her liberal pals as well as other Muslims.}  \textbf{Output:} \emph{FoxNews (vs NYTimes or WaPost)} \\ 
            \cline{2-5}
             & \makecell{summarization} & 2009-2016 & 330$k$ & \textbf{Input:} \emph{The Consumer Financial Protection Bureau is demanding PayPal return \$15 million to consumers and pay a \$10 million fine for ...}  \textbf{Output:} \emph{The CFPB alleges many customers unwittingly signed up for PayPal Credit} \\ 
             \specialrule{.001em}{.001em}{.001em}
             Food Reviews & \makecell{review rating \\ classification} & 2013-2019 & $126k$ & \textbf{Input:} \emph{What a beautiful store and amazing experience! Not only the atmosphere, but the people...}  \textbf{Output: } \emph{4 (out of 5)} \\ 
        \bottomrule
    \end{tabular}
    }
    \caption{The tasks from four domains studied in this paper, with examples. See Section~\ref{subsec:datasets} for more details. 
    }
    \label{tab:task:examples}
\end{table*}
We describe the eight tasks and four domains used for this study. Three (out of eight) of the tasks are newly defined in this work, and all tasks required nontrivial postprocessing. We provide examples and detailed statistics in Table~\ref{tab:task:examples}.

\paragraph{Domain 1: Twitter}

Social media platforms like Twitter have been mined to study aspects of language change over time, such as the introduction or diffusion of new words \cite{eisenstein2014diffusion,tamburrini2015twitter,wang2017detecting}. 
We collect unlabeled data for domain adaptation by extracting 
a  random selection of 12M tweets, spread semi-uniformly from 2015 till 2020.\footnote{Collected via the \href{https://developer.twitter.com/en/products/twitter-api/academic-research}{Twitter API}.}
We experiment with two tasks on Twitter data:

\newcommand\sbullet[1][.5]{\mathbin{\vcenter{\hbox{\scalebox{#1}{$\bullet$}}}}}
\newcommand{\taskname}[1]{\vspace{0.1cm} \noindent \emph{\underline{#1}}}

\newcommand{\poliaff}{\textsc{PoliAff}}

\taskname{Political affiliation classification (\textbf{\poliaff})}
We collect English tweets dated between 2015 and 2020 from U.S.~politicians with a political affiliation (\emph{Republican} or \emph{Democrat}).
We omit any politician who changed parties over this time period or identified as independent. 
 We consider the downstream task of detecting political affiliations, i.e., given a text of a single tweet we predict the political alignment of its author at the time of the tweet.
This task can be useful for studies that involve an understanding of ideologies conveyed in text~\cite{Lin2008AJT,iyyer2014political}.

\newcommand{\twiet}{\textsc{TwiERC}}

\taskname{Named entity type classification (\textbf{\twiet})}
We use the Twitter NER dataset from~\citet{rijhwani2020temporally}. 
The dataset contains tweets dated from 2014 to 2019, each annotated with the mentions of named entities and their types (\emph{Person}, \emph{Organization}, or \emph{Location}). 
We consider the task of typing a given mention, which is a subproblem of named entity recognition. 


\paragraph{Domain 2: Scientific Articles}
Scientific research produces vast amounts of text with great potential for language technologies
\citep{wadden2020fact,lo2020s2orc}; it is expected to show a great deal of variation over time as ideas and terminology evolve.
For adaptation to this domain, we collect unlabeled data from science documents available in Semantic Scholar's corpus,\footnote{\url{https://api.semanticscholar.org/corpus/}
} which yields 650k documents, spread over a 30-year period~\cite{ammar:18}. For this domain, we study two tasks:

\newcommand{\sci}{\textsc{SciERC}}

\taskname{Mention type classification \textbf{(\sci)}}
We use the \emph{SciERC} dataset from~\citet{luan2018multi} which contains entity-relation annotations for computer science paper abstracts for a relatively wide range of years (1980s to 2019). 
We subdivide the annotated data into time periods with roughly equal-sized numbers of papers (1980--1999, 2000--2004, 2005--2009, 2010--2016). 
The task is to map a mention of a scientific concept to a type (\emph{Task}, \emph{Method}, \emph{Metric},
\emph{Material}, \emph{Other-Scientific-Term}, or \emph{Generic}).

\newcommand{\ai}{\textsc{AIC}}

\taskname{AI venue classification (\textbf{\ai})}
We also examine \temporaldrift on the task of identifying whether a paper was published in AAAI or ICML. We group the data into roughly equal-sized time periods (2009--2011, 2012--2014, 2015--2017, and 2018--2020). 
This task is, loosely, a proxy for topic classification and author disambiguation applications~\cite{subramanian2021s2and}.

\paragraph{Domain 3: News Articles}
News articles make up a significant part of the data commonly used to train LMs~\cite{dodge2021documenting}. 
News articles convey current events, suggesting strong temporal effects on topic. 
For adaptation, we use 
9M articles from  the Newsroom dataset \cite{grusky-etal-2018-Newsroom}, ranging from 2009--2016.\footnote{\url{https://lil.nlp.cornell.edu/newsroom}}
We experiment with three tasks on news articles: 

\newcommand{\summarization}{\textsc{NewSum}}

\taskname{Newsroom summarization (\textbf{\summarization}) }
The Newsroom dataset provides a large quantity of high-quality summaries of news articles \cite{grusky-etal-2018-Newsroom}. 
We group articles by years for this task (2009--2010, 2011--2012, 2013--2014, 2015--2016). Note that this task, unlike the other tasks considered here, is not a document classification task.

\newcommand{\publisher}{\textsc{PubCls}}

\taskname{Publisher classification (\textbf{\publisher}) }
The Newsroom dataset also provides metadata, such as publication source. We take the documents published by the 3 most 
prolific
publishers (Fox News, New York Times, and Washington Post) and train models to classify documents among them. We bin the years  (2009--2010, 2011--2012, 2013--2014, 2015--2016).  This task is a proxy for applications that seek to infer fact provenance~\cite{zhang2020said}. We note that, unlike in our other tasks, we downsample to ensure that the labels are equally balanced.


\newcommand{\mediaframes}{\textsc{MFC}}

\taskname{Media frames classification (\textbf{\mediaframes})}
``Framing'' often refers to the emphasis or deemphasis of different social or cultural issues in the media's presentation of the news \cite{Entman1983FramingTC}. \citet{card-etal-2015-media} provide a dataset of news articles annotated with framing dimensions. We predict the primary frame of a document, treating the problem as a 15-way classification task. We bin by timestamp (2009--2010, 2011--2012, 2013--2014, 2015--2016).

\paragraph{Domain 4: Food Reviews}
Food and restaurant reviews have been widely studied in NLP research. We considered this domain as a possible contrast to those above, expecting less temporal change.  Using data from the Yelp Open Dataset,\footnote{\url{https://www.yelp.com/dataset}}  we consider one task:
 

\newcommand{\yelpreview}{\textsc{YelpCls}}

\taskname{Review rating classification (\textbf{\yelpreview}) }
This is a conventional sentiment analysis task, mapping the text of a review to the numerical rating given by its author \cite{Pang2002ThumbsUS, Dave2003MiningTP}.  We partition the data by year (2013 to 2019) and ensure that each timestep has a roughly equal amount of reviews.

\section{Empirical Results and Analysis}

\label{sec:empirical}

In this section, we summarize our experimental analysis, resulting from more than 500 experiments. In our experiments, we primarily explore the effect of \temporaldrift on GPT2~\cite{brown2020language}, a  LM often used for generation.\footnote{In our preliminary results, we found that BERT, RoBERTa, and GPT2 models showed similar patterns.} 
We report the macro $F_1$ score for  classification tasks and \emph{Rouge-L}~\cite{Lin2004ROUGEAP} for \summarization{}.

We first focus on quantifying \temporaldrift in end tasks. As a preliminary analysis, we investigate how the marginal distribution over labels changes over time. We then study how \temporaldrift affects performance of GPT2 models in downstream tasks with temporal finetuning (\hyperref[Q1]{\subscript{Q}{1}},\hyperref[Q2]{\subscript{Q}{2}}). We find that the amount of performance degradation can vary by task; in some cases the degradation can be severe.

We then study how \temporaldrift affects LMs. As a first step, we  analyze how vocabularies change over time in our datasets. We then experiment with \ref{Q3} how \temporaldrift affects upstream language modeling and \ref{Q4} how effective temporal adaptation, or additional pretraining on a target year, is in mitigating misalignment. We find that while LMs are affected by misalignment, temporal domain adaptation is not enough to mitigate \temporaldrift.

Details on temporal domain adaptation and finetuning, and an extended version of our results, can be found in Appendices~\ref{sec:appendixA} and \ref{sec:appendixB}, respectively.

\subsection{Temporal Misalignment in Tasks}
\label{sec:downstream:tasks}

How much does misalignment affect task performance?  We  find that it depends on the task.

\paragraph{Label Distribution Drift}
\label{sec:labeldistr}
\begin{figure}[t]
\centering
\includegraphics[width=.4\textwidth,trim=0cm 0.4cm 0cm 0.3cm]{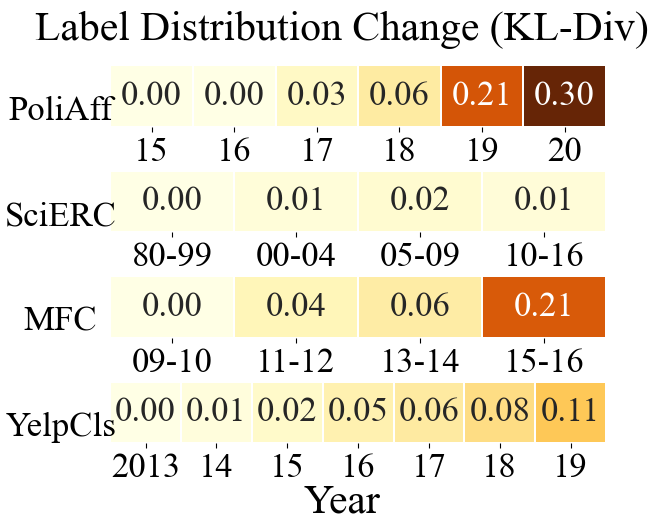}
\caption{
    KL divergence between label distributions over time for a subset of tasks. See Appendix~\ref{sec:appendixB} for full results. For each cell, we compare the distribution of labels to that of the first time period; e.g., the 2017 \poliaff{} cell contains the KL-divergence between the label distributions of \poliaff{} in 2017 and 2015. While most tasks see little change over time, \poliaff{} and \mediaframes{} see a large shift. 
}
\label{figure:label_distr_subs}
\end{figure}

We first investigate how task datasets  undergo changes in the marginal distribution over labels due to time.  For each task and each test period, we calculate the KL divergence between the label distributions in that period and the first test period. 
Full results are reported in Fig.~\ref{figure:label_distr_subs}. 
In three cases, we detected notable label distribution drift:  \poliaff{}, \ai{}, and \mediaframes{}.\footnote{For other tasks, it is possible that the data collection/annotation procedures suppressed label distribution changes that would be visible in data ``from the wild.''}
In \poliaff{}, Republican tweets outnumbered  Democratic ones by over a 2:1 ratio in 2015, but the reverse held by 2020.  This observation shows that, regardless of the properties of NLP models, the nature of many tasks changes over time, if only because the output distribution changes.

\paragraph{Finetuning}
\label{subsub:finetuning}
\begin{figure*}
\centering
\includegraphics[width=0.95\textwidth,trim=0cm .35cm 0cm 0.5cm]{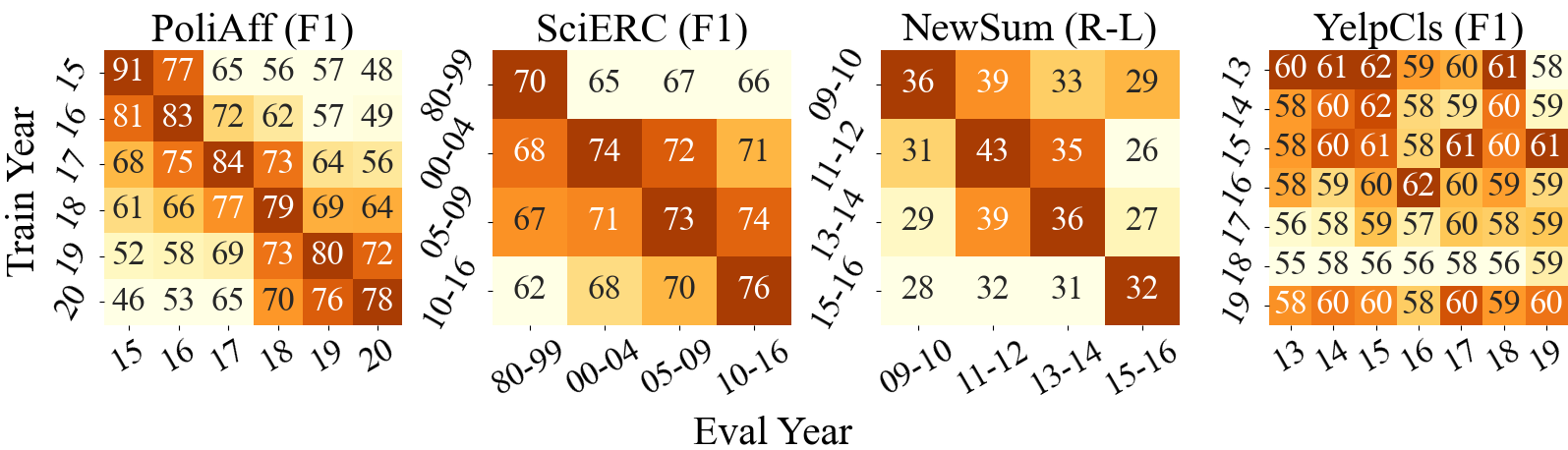}
\caption{
    Temporal misalignment in finetuning affects task performance (\S\ref{subsub:finetuning}). In all cases, higher scores are better. The heatmap is shaded per column, i.e., the darkest shade of \orangetext{orange} in a cell means the cell has the highest score in that column.
    Mismatch between the the training and evaluation data can result in massive performance drop; the degree varies by task. 
    For example, \yelpreview{} shows minimal degradation. In contrast, \poliaff{} shows major deterioration over time. Additional tables for remaining tasks can be found in Appendix \ref{sec:appendixB}.
å}
\label{table:gpt2_downstream_tasks}
\end{figure*}

As described in \S\ref{subsec:datasets}, for each  task, we create training and evaluation sets associated with different time periods.  We finetune GPT2 on each of the task's training sets and evaluate each on two evaluation sets.  Note that there is no domain adaptation here.

\begin{table}[t]
\centering
\small

\pgfplotstabletypeset[
    typeset cell/.append code={%
    \ifx#1\pfft%
    \pgfkeyssetvalue{/pgfplots/table/@cell content}{\\ \midrule}%
    \fi},
    columns={Z,A,B,C, dummy},
    col sep=comma, 
    every head row/.style={before row=\toprule, after row=\midrule },
    every last row/.style={after row=\bottomrule},
    columns/Z/.style={ reset styles, string type,
    column name={Domain}, column type={l}},
    columns/A/.style={reset styles, string type, , column name=Task (metric), column type={|l}},
    columns/B/.style={ column name=TD,fixed zerofill, column type={|l}},
    columns/C/.style={ column name=$r$, fixed zerofill},
    columns/dummy/.style={string type,column name={}}
]{table.dat}
    \caption{
        Finetuned models' temporal degradation summary scores (TD; \S\ref{subsec:metric}; details in Figure~\ref{table:gpt2_downstream_tasks}). 
        These scores estimate how fast a model degrades as the time period of training and evaluation data diverge (higher scores imply faster degradation). We note that since we normalize by the overall time range of a task, the temporal partitions we used do have an effect on the TD scores. For example, \ai{} spans ten years, even though there are only four partitions.
        \added{We also show the correlation coefficient, $r$, that measures the strength of a linear relationship (0 meaning no correlation, 1 being perfectly correlated). In all cases but Yelp, the degree of degradation has a moderate correlation with the distance between the training and evaluation years ($r>0.5$, $p<0.05$). We use the Wald test with the null hypothesis that the slope is 0. }
    }
    \label{tab:td:scores:fine-tuned} 
\end{table}

Fig.~\ref{table:gpt2_downstream_tasks} shows our results on downstream tasks (with no domain adaptation). 
To get more reliable estimates, each number in this heatmap is an average of five independent experiments with different random seeds.  
A summary of the fine-tuning results, in terms of TD scores (\S\ref{subsec:metric}) is in Table~\ref{tab:td:scores:fine-tuned} which indicates the speed of temporal degradation, for every year that the training and evaluation data diverges. Recall that this score (applied to task performance measures) summarizes the strength of the effect of temporal misalignment on the score, using evidence from across experiments.


\paragraph{\ref{Q1} Temporal misalignment degrades task performance substantially.
}
Fig.~\ref{table:gpt2_downstream_tasks}, similar to earlier work~\cite{rottger2021temporal}, shows that models trained on data from the same time period as the test data perform far better than those from the past.  The performance drop is most severe for  \poliaff{} (TD=$7.72$) and \publisher{} (TD=$5.45$).


\paragraph{\ref{Q1} Temporal misalignment has a measurable effect on most tasks.}
Half of our tasks see an average loss of at least 1 point for each time period that the training data diverges from the test data. For datasets like \sci{} that make use of data from three decades or more, this effect could add up.

Moreover, 1 point of difference can be meaningful, especially for the summarization task where we measure {Rouge-L}. According to the leaderboard,\footnote{\url{https://lil.nlp.cornell.edu/newsroom/index.html}} the best three performing models are within a point of each other in {Rouge-L}~\cite{shi-etal-2019-leafnats, Shi2021NeuralAT, Mendes2019JointlyEA}. The task has a TD score of 1.38. On average, a time period of \temporaldrift results has a larger effect on performance than changing between the three best models.


\paragraph{\ref{Q1} Performance loss from temporal misalignment occurs in both directions.}
Another  observation in Fig.~\ref{table:gpt2_downstream_tasks} is that degradation happens in both directions (past and future).
While most of the emphasis on \temporaldrift is on how to adapt our stale models/data to the present time~\cite{DhingraTemp, lazaridou2021pitfalls, rottger2021temporal}, our experiments also show that models trained on newer data  can be misaligned from the past, as well. \added{ Weak performance in older texts has been noted in NLP for historical documents \cite{Yang2016PartofSpeechTF,Han2019UnsupervisedDA}. However, our findings indicate deterioration can occur sooner---just a few years rather than decades or centuries.} \kelvin{idk I can also delete this paragraph.} \nascomment{keep (I edited slightly)}\daniel{+1 for keeping it}

\paragraph{\ref{Q2} Tasks, even in the same domain, are affected differently.} 
Consider the two tasks of \poliaff{} and \twiet{} (both in the Twitter domain), with TD scores of 7.72 and 0.96, respectively. 
Of our 8 tasks, \twiet{}, \mediaframes{}, and \yelpreview{} are the most robust to temporal misalignment (TD scores of 0.96, 0.98 and 0.26, respectively).  The high levels of variation show that temporal misalignment  affects performance through labeled datasets, not just unlabeled pretraining data.

\subsection{Temporal Misalignment in LMs}

As LMs are widely used in modern NLP systems, it is important to inspect how robust they are to \temporaldrift. We seek to understand how \temporaldrift affects the language modeling task in our four domains and if temporal domain adaptation helps in downstream tasks.

\paragraph{Vocabulary Shift}
\label{sec:vocabshift}
\begin{figure*}[t]
\centering
\includegraphics[trim=0cm 0.6cm 0cm 0cm, width=\textwidth]{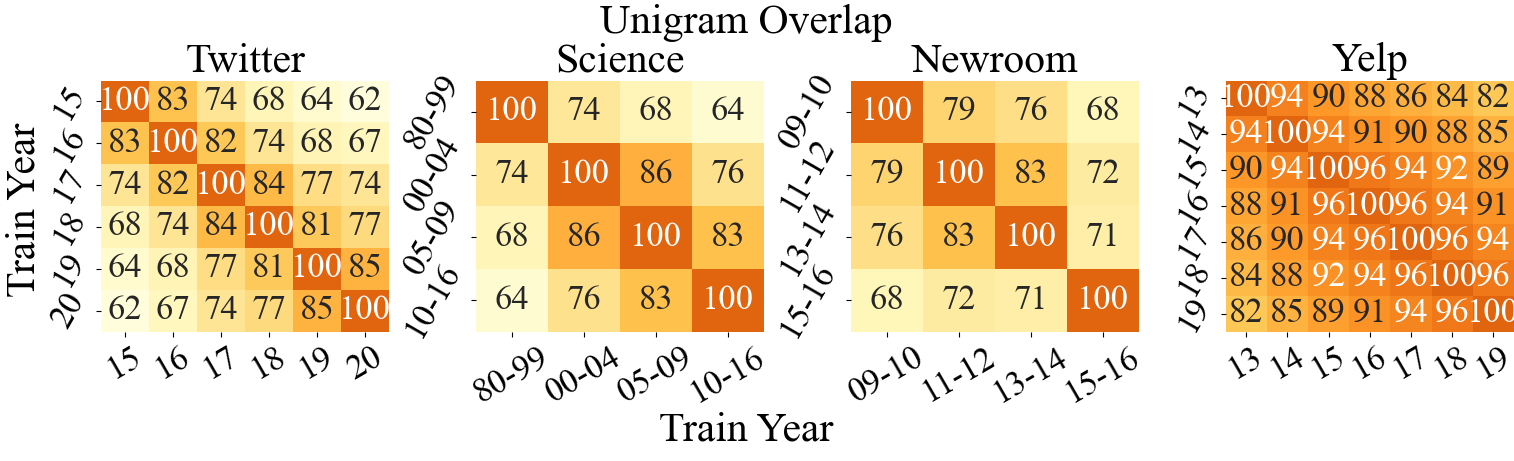}
\caption{
    Vocabulary overlap between time periods, over a subset of our tasks' datasets. Each cell shows the \% overlap between the vocabularies of two time periods.} 
\label{figure:word_overlap}
\end{figure*}

We first consider an extremely simple measurement of language shift:  how do vocabularies change across time periods?\footnote{This can be understood as a model-free way to measure covariate shift for NLP tasks that take text as input.} We use a similar procedure to the one \citet{gururangan2020don} used for analyzing domain similarity. Fixing a domain, we compare the (unigram) vocabularies of each pair of training sets.
The vocabularies are built using the 10K most frequent terms from each time period.
 We note that vocabulary overlap is higher between two time periods the closer they are.  Most domains see a sizeable amount of shift; however, Yelp is relatively  stagnant. Fig.~\ref{figure:word_overlap} visualizes the overlap measurement. \added{Table~\ref{tab:overlapcorr} in Appendix~\ref{sec:appendixB} shows the correlation between model performance and the word overlap.} \kelvin{not married to this table; can remove or shove to appendix too.} \nascomment{leave it} 


\paragraph{Temporal Domain Adaptation}
\label{sec:domain:adaptation}

\begin{figure}[t!]
\centering
\includegraphics[width=.38\textwidth,trim=1.9cm .5cm 1.0cm 0.9cm]{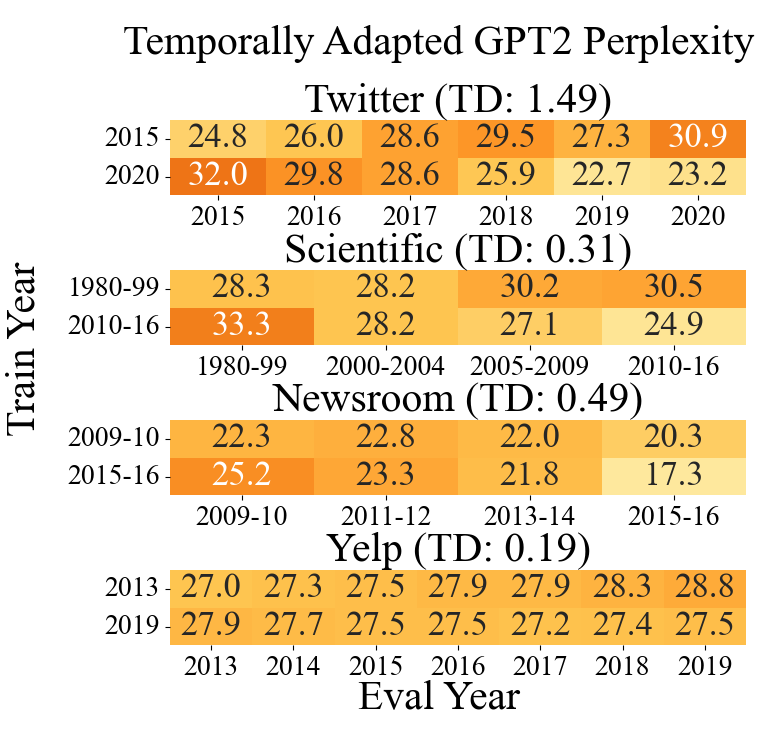}
\caption{
    Perplexity of GPT2 after adaptive pretraining on temporally-selected data in different domains (lower is better). The TD score (in parentheses) estimates the expected perplexity rise (i.e., degradation) for every time period of misalignment between evaluation and training times.  Degradation follows the expected pattern, but the magnitude varies by domain. 
}
\label{figure:gpt2_lm}
\end{figure}

\added{Researchers have studied the broader problem of distributional shift \cite{Shimodaira2000ImprovingPI,Zhang2013DomainAU}. The NLP community has historically tackled these problems via domain adaptation \cite{Jiang2007InstanceWF, daume-iii-2007-frustratingly, gururangan2020don}. Taking inspiration from these approaches,} we next apply DAPT to GPT2, \added{treating each time period as a domain}:  for each time period, we continue pretraining and then evaluate perplexity.  We consider how the perplexity varies with the (mis)alignment between the DAPT training data and the evaluation data. We measure the TD score, which summarizes how much performance is affected by \temporaldrift (now applied to perplexity). The results of  temporal domain adaptation are  in Fig.~\ref{figure:gpt2_lm}. 

\paragraph{\ref{Q3}  Domains are a major driver of \temporaldrift in LMs.}
Consistent with \citet{lazaridou2021pitfalls},   Fig.~\ref{figure:gpt2_lm} shows degradation of LM due to \temporaldrift; it further shows considerable variation by text domain.  Twitter changes most rapidly, and food reviews are much slower.  This observation is consistent with past work on language change in social media  \cite{stewart-eisenstein-2018-making, eisenstein2014diffusion}.
To the extent that a LM's practical usefulness is associated with its fit to new data, researchers and practitioners should understand the temporal dynamics of their target text domains and plan LM updates accordingly.

\paragraph{Joint Effects of Temporal Adaptation and Finetuning }
\label{subsub:adaptaton:plus:tuning}

As discussed in \S\ref{sec:setup}, continued pretraining of an LM on in-domain text has been shown to improve task performance. Our prior results show that both downstream tasks and language modeling are affected by \temporaldrift. Can temporal domain adaptation help mitigate the effects of misalignment in downstream tasks?

Here we consider how the time period of the data selected for continued pretraining affects task performance.
For each task's evaluation set, we apply DAPT twice:  once with the earliest available time period's unannotated data and once with the latest's.  We then finetune and evaluate on data from the same time periods as in the earlier experiment. 


\paragraph{ \ref{Q4}
Temporal adaptation does \underline{not} overcome degradation from temporally misaligned labeled data.
}
In Table~\ref{table:gpt2_dapt_small},
we see small performance gains from temporal domain adaptation on LMs, and in some cases it is harmful.  These observations underscore the importance of the labeled data; adjustments to the LM alone do not yet appear sufficient to mitigate the effects of temporal misalignment.
In contrast to temporal domain adaptation, which does not mitigate temporal misalignment's effects,  finetuning on temporally-updated labeled data is more effective.

This can be observed in each task-specific sub-table of in Table~\ref{table:gpt2_dapt_small}: 
the top-left and bottom-right quadrants (fine-tuning on time-stamp that is used for evaluation) generally lead to higher scores. 
\begin{table*}[]
    \centering
    \includegraphics[width=0.99\linewidth,trim=4.34cm 31.2cm 4.1cm 1.85cm,clip=true]{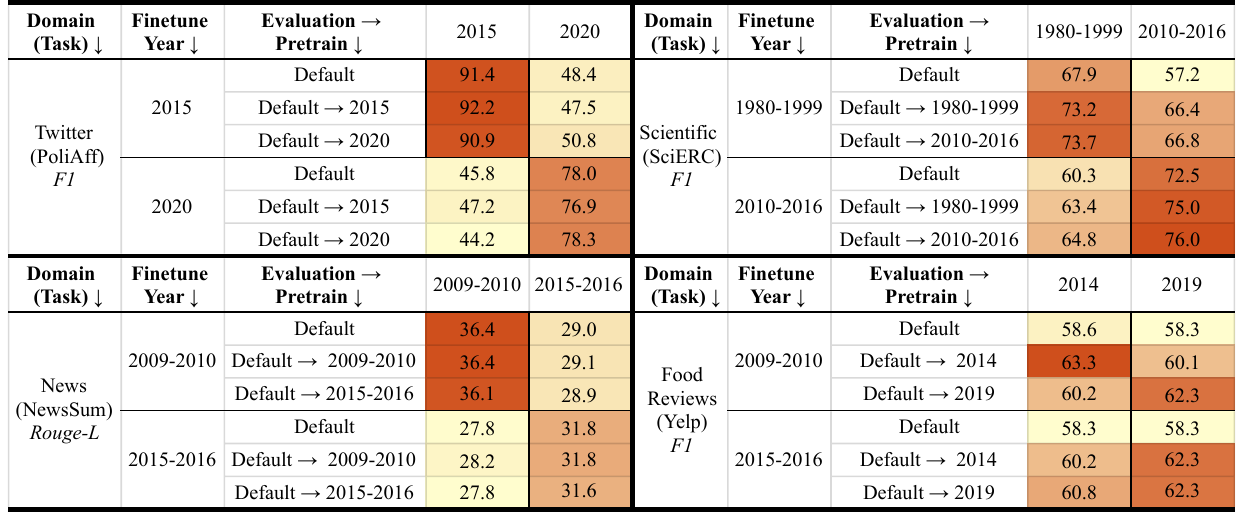}
    \caption{
        Combination of temporal adaptation and finetuning  (\S\ref{subsub:adaptaton:plus:tuning})  on our tasks. 
        The row labeled ``Default'' corresponds to a model that has not been adapted (uses the default pretraining). The models with temporal domain adaptation are shown in rows labeled ``Default $\rightarrow y$'' and each is comparable to the ``Default'' row above it.
        The color coding is proportional to the magnitude of the performances of each task (darker shade of \orangetext{orange} indicates higher scores). 
        It can be observed that temporal finetuning has a greater impact than temporal pretraining. Each quadrant of 3 for each task, indicating the same finetune and evaluation years, but different pretraining conditions, are mostly uniform. In contrast, we notice a sharper difference in performance when varying the finetuning year (comparing the quadrants vertically). \kelvin{Asking for opinions: swap FT and Pretrain columns or keep as is?}  \nascomment{no opinion} \suchin{If it's not too hard, I would swap, since pretrain comes before fine-tuning, and it's more intuitive to read it from left to right.}
    }
\label{table:gpt2_dapt_small}
\end{table*}
\section{Limitations and Future Work}

We provided a well-controlled suite of experiments to study the effects of temporal misalignment on model performance. However, the setup has some drawbacks. For example, we expect that models trained on data accumulated across multiple time periods would perform well \cite{lazaridou2021pitfalls, rottger2021temporal,lifelongpretrainingadapt}. 

We chose the time periods in our study so that they would each have sufficient and consistent training data sizes. However, amounts of data in a particular domain or task will fluctuate over time. Moreover, the rate of language use change may not be uniform. Time periods should be selected with these two considerations in mind.

Our findings indicate that temporal misalignment's effects depend heavily on the task.  Though not studied here, the same issues may arise in annotation efforts; consider, for example, recent work on controversy \cite{Zhang+al:18a} and social norms \citep{xu2021detoxifying, zhou2021challenges} likely hinges on constructs that may be time sensitive.  Annotations that are temporally misaligned with the original data being annotated may be anachronistic.

An opportunity for future exploration is in the context of real-world events with sudden changes such as COVID-19 pandemic~\citep{cao-etal-2021-quantifying} or political changes, which influence tasks such as question answering~\citep{DhingraTemp, zhang2021situatedqa}. 

\added{ Extensive work has been done on modeling and detecting lexical semantic change, or how words evolve in meaning \cite{Hamilton2016DiachronicWE, Rudolph2018DynamicEF, Gonen2020SimpleIA}. Techniques and intuition from this body of work may be useful in finding solutions to mitigate degradation due to misalignment. We believe that this phenomenon is an important aspect of temporal misalignment, but leave disentangling semantic shifts from other, perhaps task-related factors, for future work. }


Continual learning, which allows models to learn from a continuous stream of data, could also be one way to mitigate \temporaldrift{}. Most prior work in this space has focused on continual learning in LMs \cite{lifelongpretrainingadapt} or learning disparate tasks \cite{NEURIPS2019_f8d2e80c, huang-etal-2021-continual}. Future work may investigate continual learning algorithms for tasks that change over time.

\added{Our results showed that straightforward domain adaptation was unable to mitigate the effects of temporal misalignment. Recent work in language modeling has elevated the importance of domains by using a mixture of domains \cite{demix} or giving domains a hierarchical structure \cite{Chronopoulou2021EfficientHD}. More sophisticated approach to domains, in line with these works, could lead to temporally robust models.}

While we found that task-specific finetuning is more effective than temporal adaptation, new labeled data can be expensive. Ways to characterize or detect changes in a task could be helpful in efficiently updating datasets \cite{Lu2019LearningUC,Webb2018AnalyzingCD}. Future work can also treat dataset maintenance as an optimization problem between the cost and gains of annotating new data~\cite{bai2021pretrain}. 

\section{Conclusion}

Changes in language use over time, and how language relates to other quantities of interest in NLP applications, has clear effects on the performance of those applications.  We have explored how temporal misalignment between training data---both data used to train LMs and annotated data used to finetune them---affects performance across a range of NLP tasks and domains, taking advantage of datasets where timestamps are available. We compile these datasets as a benchmark for future research as well.
We also introduced a summary metric, TD score, that makes it easier to compare models in terms of their \temporaldrift.  

Our experiments revealed considerable variation in temporal degradation accross tasks, more so than found in previous studies \citep{rottger2021temporal}.  These findings motivate continued study of temporal misalignment across applications of NLP, its consideration in benchmark evaluations,\footnote{Indeed, for benchmarks where training and testing data \emph{are} aligned, our findings suggest that measures of performance may be in some cases inflated.} and vigilance on the part of practitioners able to monitor live system performance over time.

Notably, we observed that continued training of LMs on temporally aligned data does not have much effect, motivating further research to find effective temporal adaptation methods that are less costly than ongoing collection of annotated/labeled datasets over time.

\section*{Acknowledgments}

\added{We thank Dallas Card, Sihao Chen, Alexander Fabbri, Jack Hessel, and Shruti Rijhwani for their help in curating our data. We thank Jacob Eisenstein, Rahul Nadkarni, the ARK and XLab research groups, and our anonymous reviewers for their feedback on this work. We also acknowledge the Beaker team (\url{https://beaker.org}) for their support with experiments.} This research was supported in part by the Office of Naval Research
under MURI grant N00014-18-1-2670.

\bibliography{ref}
\bibliographystyle{acl_natbib}

\clearpage

\appendix

{\Large \bf Supplementary Material }

\section{A Metric for Temporal Degradation}
\label{appendix:subsec:metric}


Let $t$ be the time period of the training data and $t'$ the time period of the evaluation data.\footnote{See examples in Fig.~\ref{table:gpt2_downstream_tasks}.} We aim to summarize the general effect of temporal misalignment (the difference between $t$ and $t'$) on task performance, in an interpretable way that is comparable across tasks.

Let $S_{t' \shortto t}$ indicate the performance a model trained on timestamp $t'$ data and evaluated on the timestamp $t$. Let 
$$
{D} (t' \shortto{} t) = 
    -\left(S_{t' \shortto t} - S_{t \shortto t} \right) \times \text{sign}(t' - t), 
$$

In other words, ${D} (t' \shortto{} t)$ is a modified difference in performance between a aligned and misaligned models. The modification ensures that, as performance deteriorates, $D$ increases, regardless of the direction of time between $t$ and $t'$.

Our temporal degradation (TD) score for a fixed evaluation timestamp $t$ for models trained on a set of timestamps $\timestamsps$ is defined as:
$$
\text{TD} (\timestamsps \shortto{} t') = \left| \frac{ \sum_{t \in \timestamsps } \left({D} (t' \shortto{} t)  - \bar{D} \right)(t - \bar{t}) }{ \sum_{t \in \timestamsps} (t - \bar{t})^2  } \right|,
$$
where $ \bar{t} = \text{avg}_{t \in \timestamsps} t' $ and $\bar{D} = \text{avg}_{t \in \timestamsps} {D} (t' \shortto{} t) $. 
This metric is  the \emph{slope} of a line fitting the the performance change of models trained on a variety of timestamps, when evaluated on a fixed timestamp.
It can be interpreted as the average rate of performance deterioration per time period.

\begin{figure*}
\centering
\includegraphics[trim=0cm 0.6cm 0cm 0cm, width=\textwidth]{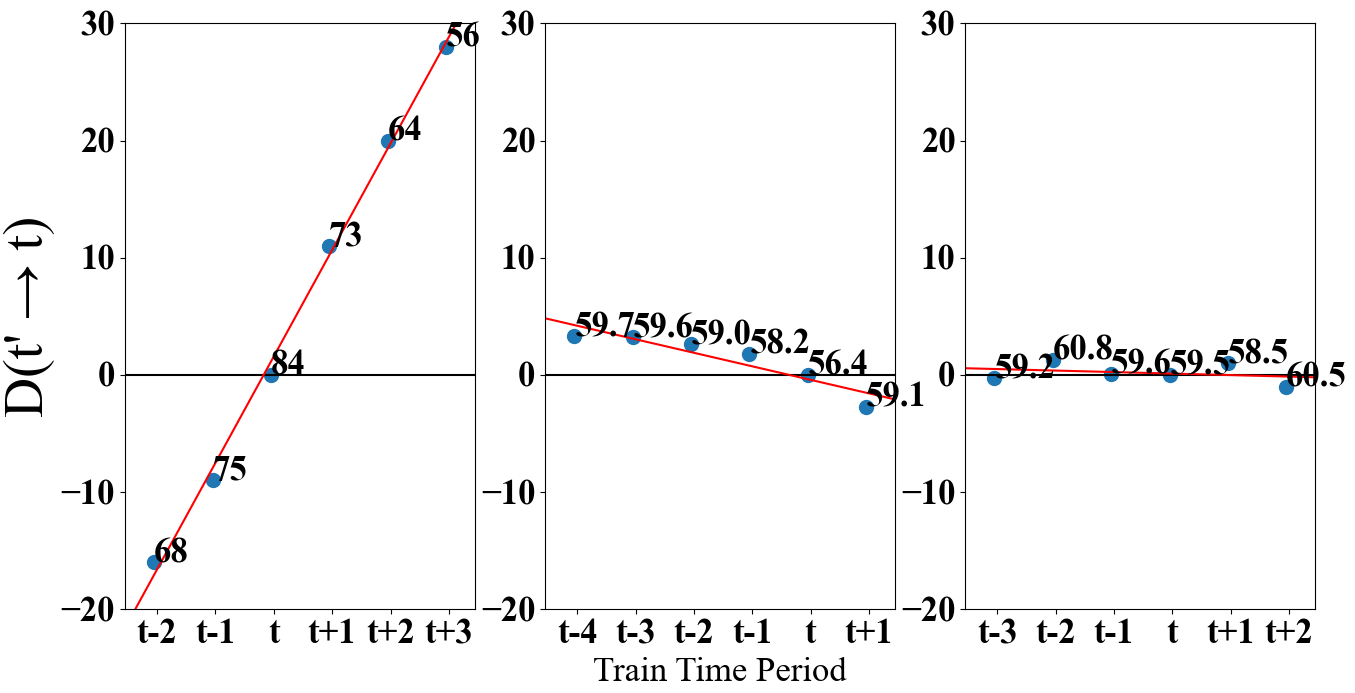}
\caption{
    Three example calculations of the TD score (left from \poliaff{} and the center and right from \yelpreview{}). The annotated numbers are the raw evaluation scores $S_{t'\to t}$ and the plotted markers represent the modified differences $D(t'\to t)$ discussed in Section~\ref{subsec:metric}. For a particular plot, the red line is the line of best fit and its slope is the $\text{TD}(t)$ score for evaluation timestep $t$. The final TD score is averaged between all evaluation timesteps for the particular task.
}
\label{figure:td_more}
\end{figure*}

Fig.~\ref{figure:td_more} shows three examples of  TD scores from \poliaff~(the first) and \yelpreview~(the latter two). These illustrate cases with and without temporal sensitivity. In practice, most examples with deterioration showed a linear trend and thus the rate of degradation was suitible to be approximated by a line. The final TD score is averaged over all evaluation years $\timestamsps'$.

$$\text{TD} = \frac{\sum_{t\in\timestamsps'}\text{TD}(\timestamsps \shortto t)}{|\timestamsps'|}$$

\nascomment{changed denominator above from $n$ which wasn't defined anywhere}

\section{Details of Model Development}
\label{sec:appendixA}

\paragraph{Training Details for Temporal Adaptation}

We train GPT2 over each domain and timestamp for $k$ steps using Huggingface's implementation of GPT2.  Hyperparameter details can be seen in Table~\ref{tab:hyperparameters:pretrain}.

\begin{table}[]
    \small
    \centering
    \begin{tabular}{@{}ll@{}}
    \toprule
     \textbf{Hyperparameter} & \textbf{DAPT Assignment}  \\ \midrule
        Number of steps & 10k\ \\ \midrule
        Batch size & 32 \\ \midrule
        Maximum learning rate & 5e-05 \\ \midrule
        Adam Epsilon & 1e-08  \\ \midrule
        Adam Beta & 0.9. 0.999  \\ \midrule
        Block size & 1024  \\ \midrule
    \end{tabular}
    \caption{Hyperparameters for temporal adaptation accross the four domains.}
    \label{tab:hyperparameters:pretrain}
\end{table}

\paragraph{Training Details for Temporal Finetuning}

We use Huggingface's implementation of GPT2 for finetuning for both the classification and summarization tasks. We train on Quadro RTX 800 GPUs. See Table~\ref{tab:hyperparameters:finetune} for details.

\begin{table}[]
    \centering
    \small
    \begin{tabular}{@{}lll@{}}
    \toprule
     \textbf{Hyperparameter} & \textbf{Cls. Assign} & \textbf{Summ. Assign}  \\ \midrule
        Number of Epochs & 50 & 10\ \\ \midrule
        Batch size & 32 & 8 \\ \midrule
        Max learning rate & 2e-05 & 2e-05\\ \midrule
        Adam Epsilon & 1e-08   & 1e-08 \\ \midrule
        Adam Beta & 0.9. 0.999 & 0.9. 0.999 \\ \midrule
        top p (sampling) & - & 0.05 \\ \midrule
        top k & - & 20 \\ \midrule
        temperature & - & 1 \\ \midrule
        max length & - & 512 \\ \midrule
                
    \end{tabular}
    \caption{Hyperparameters for temporal finetuning accross the eight tasks.}
    \label{tab:hyperparameters:finetune}
\end{table}

\section{Data Collection}
\label{sec:appendix:data}

We describe the postprocessing and data collection in greater detail. All data released is intended for non-commercial use.



\paragraph{\poliaff}
We acquire a list of U.S. politician names and Twitter handles.\footnote{\url{https://files.pushshift.io/twitter/US_PoliticalTweets.tar.gz}} One of the authors manually annotated whether each politician was a Republican or Democrat. In addition, one volunteer double checked to ensure correctness. We discard any politician who changed parties between 2015 and 2020, any independents, and anyone suspended by Twitter (e.g., \@RealDonaldTrump).

\paragraph{\ai} We randomly sample science documents in Semantic Scholar's corpus.\footnote{\url{https://api.semanticscholar.org/corpus/}; licensed under an ODC-BY} Of those, we only keep documents that (1) are published in ICML or AAAI, (2) are classified as `computer science' documents, and (3) have an abstract of at least 50 tokens.

\paragraph{Newsroom} The following applies to the postprocessing and data selection for both supervised temporal finetuning and unsupervised temporal adaptation of \publisher{} and \summarization{}. We use the Newsroom dataset.\footnote{\url{https://lil.nlp.cornell.edu/newsroom/}}. We only keep articles where (1) the year in the metadata also appears in the main text and (2) no future year is mentioned in the main text.

\paragraph{\publisher} We carry out additional postprocessing and ensure that each of the three labels (Fox News, New York Times, and Washington Post) have an equal distribution across years. We do so by uniform-random downsampling.

\section{Extended Results}
\label{sec:appendixB}

We provide further results from our experiments described in Section~\ref{sec:empirical}.


\begin{table}
\centering
\small

\pgfplotstabletypeset[
    typeset cell/.append code={%
    \ifx#1\pfft%
    \pgfkeyssetvalue{/pgfplots/table/@cell content}{\\ \midrule}%
    \fi},
    columns={Z,A,B,dummy},
    col sep=comma, 
    every head row/.style={before row=\toprule, after row=\midrule },
    every last row/.style={after row=\bottomrule},
    columns/Z/.style={ reset styles, string type,
    column name={Domain}, column type={l}},
    columns/A/.style={reset styles, string type, , column name=Task (metric), column type={|l}},
    columns/B/.style={ column name=Pearson's $r$,fixed zerofill, column type={|l}},
        columns/C/.style={ column name=$r$, fixed zerofill},
    columns/dummy/.style={string type,column name={}}
]{table2.dat}
    \caption{
    \added{        Pearson $r$ correlation coeffecients between the word overlap and performance of each task. }
    }
    \label{tab:overlapcorr} 
\end{table}

\paragraph{Word Overlap Correlation with Performance} In addition to measuring vocabularies' change over time in Section ~\ref{sec:vocabshift}, we find correlations between the word overlap and model performance of each task in Table~\ref{tab:overlapcorr}.

\begin{figure*}
\centering
\includegraphics[width=\textwidth]{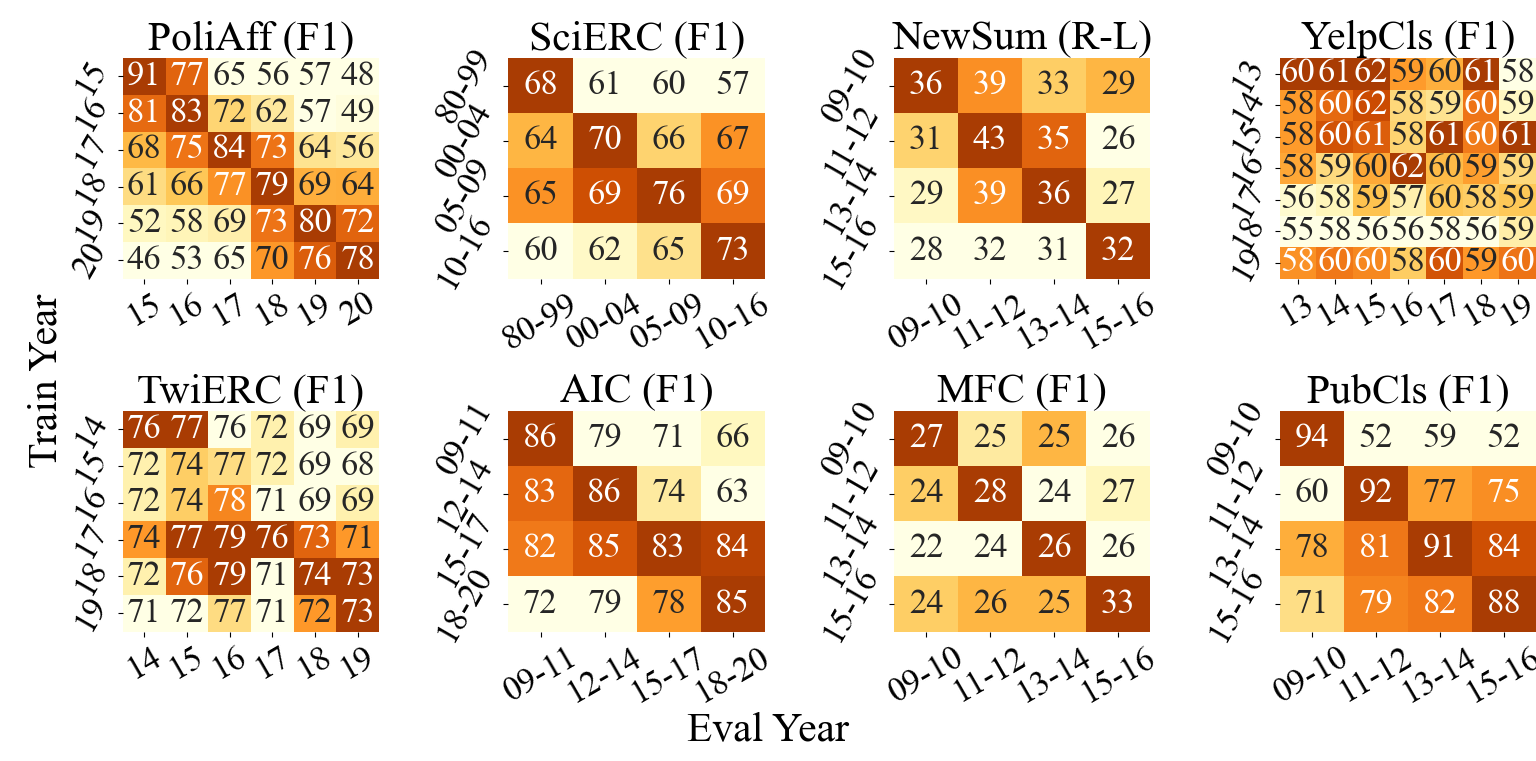}
\caption{
    Temporal misalignment in finetuning affects task performance (\S\ref{subsub:finetuning}). In all cases, higher scores are better. The heatmap is shaded per column, i.e., the darkest shade of \orangetext{orange} in a cell means the cell has the highest score in that column.
    Mismatch between the the training and evaluation data can result in massive performance drop; degree varies by task.
For example, \yelpreview{}, \mediaframes{}, and \twiet{} show minimal degradation. In contrast, \poliaff{} and \summarization{} major deterioration over time. 
}
\label{figure:gpt2_downstream_tasks_all}
\end{figure*}

\paragraph{Finetuning Results} We provide the full results from our fientuning experiments in Section~\ref{subsub:finetuning} in Fig.~\ref{figure:gpt2_downstream_tasks_all}. These results are for downstream tasks with no domain adaptation.

\paragraph{Finetuning with Temporal Domain Adaptation} We provide the full results from our finetuning with temporal domain adaptation in Section~\ref{subsub:adaptaton:plus:tuning} in Fig.~\ref{table:gpt2_dapt}.
\begin{table*}[]
    \centering
\includegraphics[width=0.99\textwidth,trim=0.8cm 22.6cm 1.5cm 1.5cm]{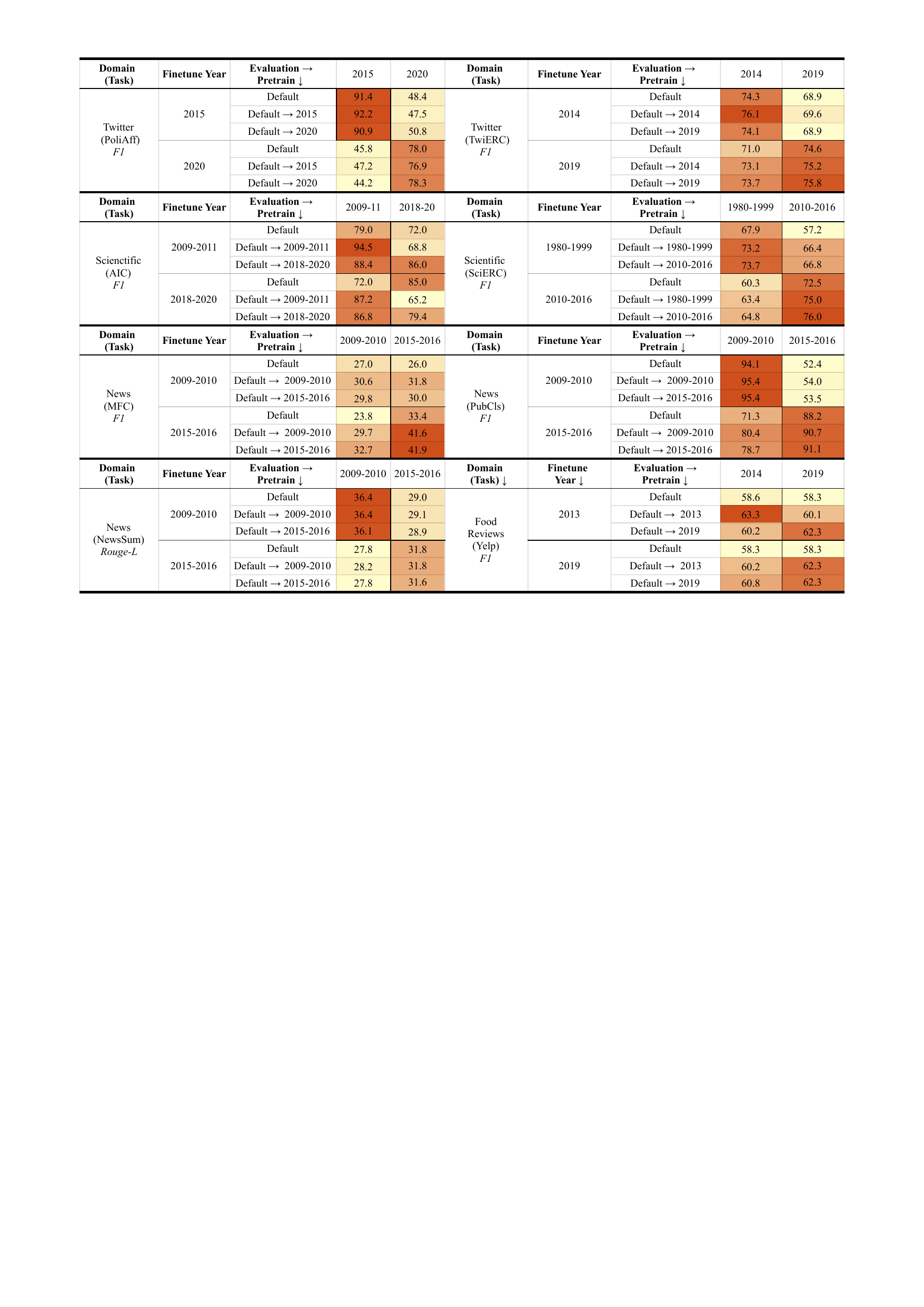}
\caption{Combination of temporal adaptation and finetuning  (\S\ref{subsub:adaptaton:plus:tuning})  on our tasks. 
The row labeled ``Default'' corresponds to a model that has not been adapted (uses the default pretraining). 
The color coding is proportional to the magnitude of the performances of each task (darker shade of \orangetext{orange} indicates higher scores). 
We see that models that were finetuned on similar time periods performed similarly, no matter how their DAPT conditions differed.
}
\label{table:gpt2_dapt}
\end{table*}

\end{document}